\documentclass[10pt,twocolumn,letterpaper]{article}

\usepackage[pagenumbers]{cvpr} 

\usepackage{graphicx}
\usepackage{amsmath}
\usepackage{amssymb}
\usepackage{booktabs}

\usepackage{subcaption,ragged2e}
\usepackage{dirtytalk}

\usepackage{algorithm}
\usepackage{algpseudocode}

\DeclareMathOperator{\EX}{\mathbb{E}}

\usepackage[pagebackref,breaklinks,colorlinks]{hyperref}

\usepackage[capitalize]{cleveref}
\crefname{section}{Sec.}{Secs.}
\Crefname{section}{Section}{Sections}
\Crefname{table}{Table}{Tables}
\crefname{table}{Tab.}{Tabs.}

\def\cvprPaperID{9221} 
\def\confName{CVPR}
\def\confYear{2022}

\begin{document}

\title{On visual self-supervision and its effect on model robustness }

\author{Michal Kucer, Diane Oyen, Garrett Kenyon\\
Los Alamos National Laboratory\\
Los Alamos, NM\\
{\tt\small \{michal, doyen, gkenyon\}@lanl.gov}
}
\maketitle

\begin{abstract}
    Recent self-supervision methods have found success in learning feature representations that could rival ones from full supervision, and have been shown to be beneficial to the model in several ways: for example improving models robustness and out-of-distribution detection. In our paper, we conduct an empirical study to understand more precisely in what way can self-supervised learning - as a pre-training technique or part of adversarial training - affects model robustness to $l_2$ and $l_{\infty}$ adversarial perturbations and natural image corruptions. Self-supervision can indeed improve model robustness, however it turns out the devil is in the details. If one simply adds self-supervision loss in tandem with adversarial training, then one sees improvement in accuracy of the model when evaluated with adversarial perturbations smaller or comparable to the value of $\epsilon_{train}$ that the robust model is trained with. However, if one observes the accuracy for $\epsilon_{test} \ge \epsilon_{train}$, the model accuracy drops. In fact, the larger the weight of the supervision loss, the larger the drop in performance, i.e. harming the robustness of the model.  We identify primary ways in which self-supervision can be added to adversarial training, and observe that using a self-supervised loss to optimize both network parameters and find adversarial examples leads to the strongest improvement in model robustness, as this can be viewed as a form of ensemble adversarial training. Although self-supervised pre-training yields benefits in improving adversarial training as compared to random weight initialization, we observe no benefit in model robustness or accuracy if self-supervision is incorporated into adversarial training.
\end{abstract}

\section{Introduction}
\label{sec:intro}

Convolutional Neural Networks are notoriously data \say{hungry} and require a large amount of labeled data to achieve adequate performance; though much work has been devoted to exploring and devising methods that do not require labeled data, or require a much smaller number of labels to learn powerful representations - e.g. unsupervised learning \cite{erhan10, kingma2013auto} and self-supervised learning \cite{rotnet, jigsaw, swav}. Self-supervised learning (SSL) has gained popularity as it proposes techniques to leverage a particular structure of the data to learn very good representations through optimizing the network weight via auxiliary tasks: e.g. in-painting \cite{inpainting}, rotation prediction \cite{rotnet} and image colorization \cite{zhang2016colorful}.

\begin{figure}[t]
  \centering
   \includegraphics[width=1.00\linewidth]{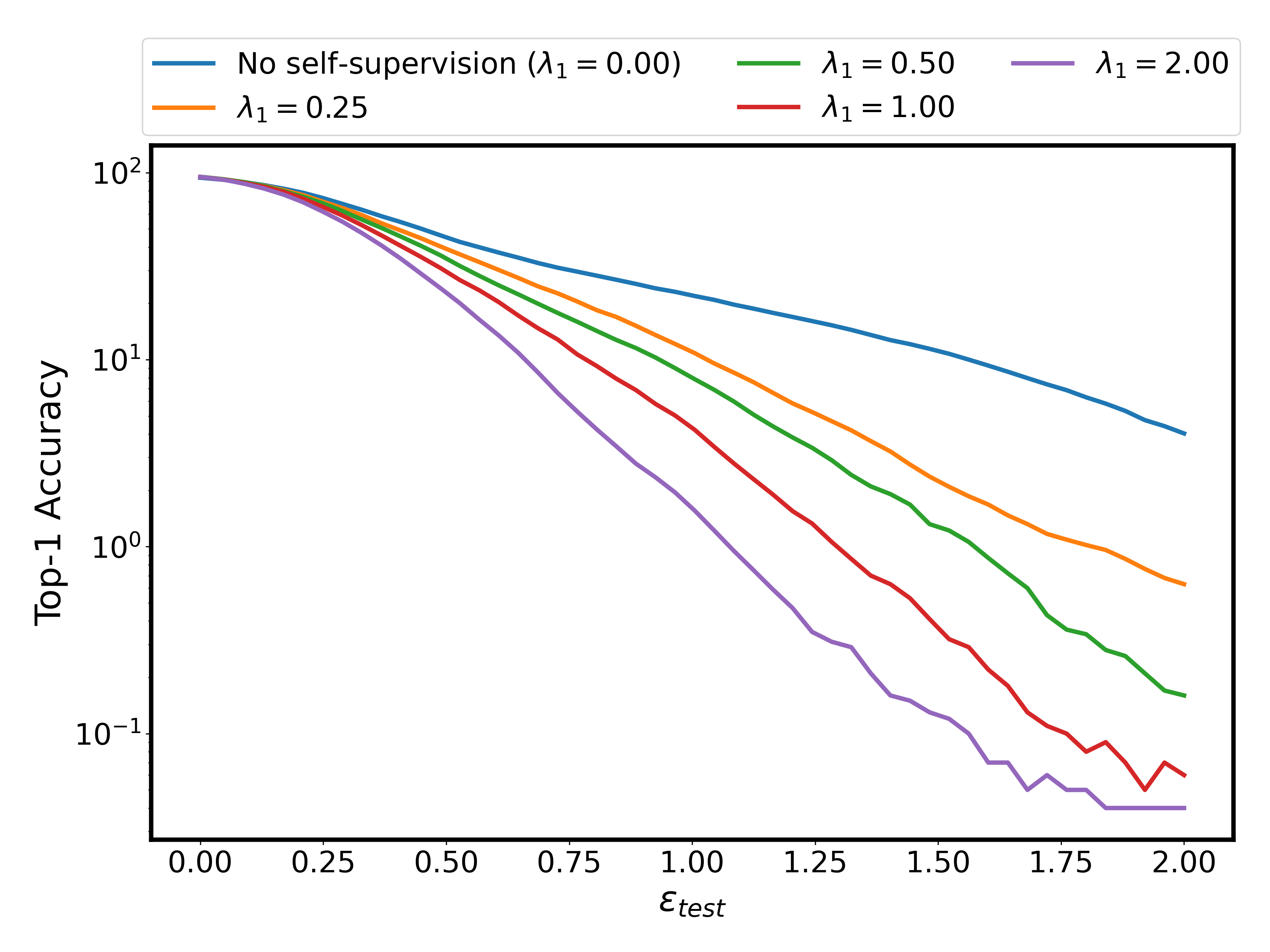}
   \caption{Example showing that increasing $\lambda_1$, the weight of self-supervised loss when combined with supervised loss, affects performance of the ResNet-18 \cite{he2016deep} architecture as $\epsilon_{test}$ is varied. Model trained with maximum allowable $l_2$ perturbations of $\epsilon_{train} = 0.10$.
   }
   \label{fig:intro}
\end{figure}

\begin{figure*}[t]
  \centering
   \includegraphics[width=0.95\linewidth]{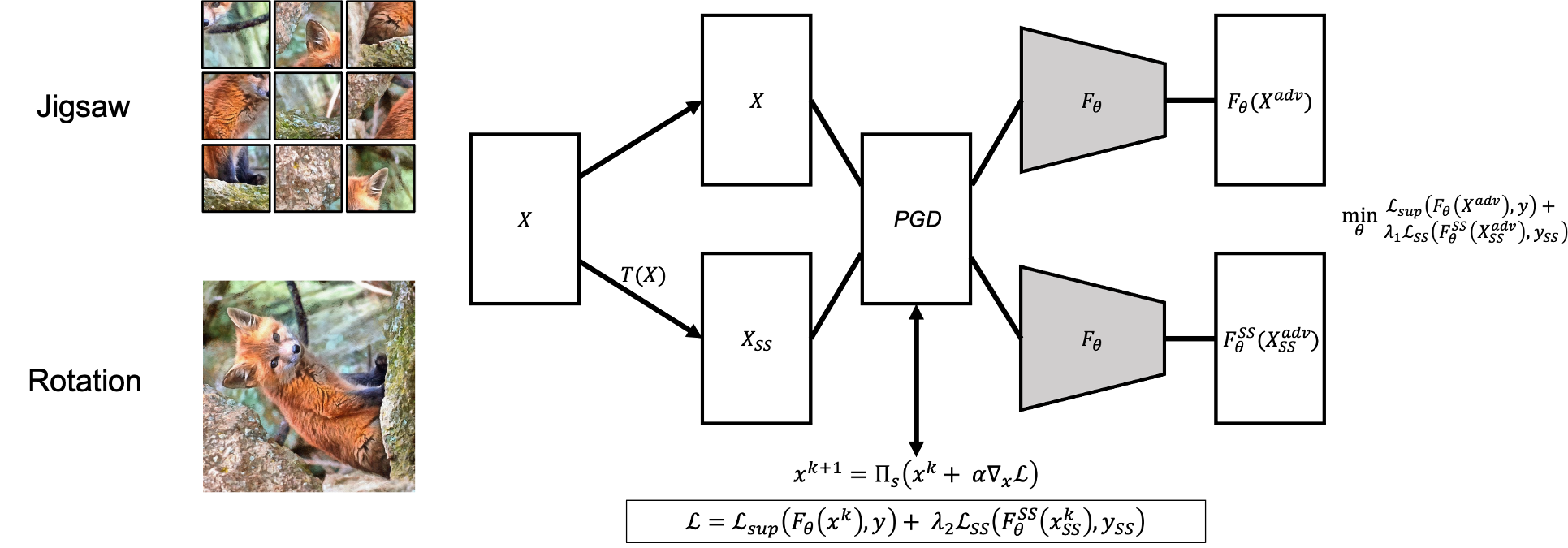}
   \caption{Figure showing the the two types of self-supervised tasks we studied and a general diagram of our model, in which we highlight where a self-supervised loss can be utilized - while generating adversarial examples, or optimizing network weights.}
   \label{fig:diagram}
\end{figure*}

Though neural networks achieve state of the art results in many tasks and domains, much recent work has focused on defending against and understanding adversarial examples, one of the main weaknesses of deep models \cite{szegedy2013intriguing, goodfellow15, wong2018provable}. Adversarial examples are inputs which have been imperceptibly changed and cause a model to provide erroneous output. However, recent work \cite{sslRobustness, sslHendrycks} suggests that leveraging self-supervised learning can lead to improvements in robustness of models. The goal of this paper is to understand how does incorporating self-supervision into model training affect model robustness to adversarial perturbations \cite{madry2018towards} and common image corruptions \cite{hendrycks2018benchmarking}, as there is more to using self-supervision to improve robustness than meets the eye. Figure \ref{fig:intro} shows a plot of the performance of an adversarially trained ResNet-18 \cite{he2016deep} model on the CIFAR-10 dataset ( with $l_2$ constraint and $\epsilon  = 0.1$) when a self-supervision (SS) loss is also used in network parameter optimization. We can see from this figure, that as we increase the maximum allowable perturbation $\epsilon$ of the adversarial examples, the model robustness worsens as the importance of SS loss, $\lambda_1$, increases. Training a model that optimizes both SS loss and supervised loss with larger $\epsilon_{train}$ further reveals that 
\begin{itemize}
    \item for several value of $\epsilon_{test} \le \epsilon_{train}$, the accuracy of a model trained with SS loss in tandem with supervision is higher, and thus one can say the model is more robust at that $\epsilon_{test}$; and,
    \item if $\epsilon_{test} \ge \epsilon_{train}$, the performance of the model trained with self-supervision drops, and the effect is larger with stronger weight $\lambda_1$ of the self-supervised loss.
\end{itemize}

This represents only one of the cases in which self-supervision can be added to adversarial training. Refer to Figure \ref{fig:diagram} for a diagram that shows ways that self-supervised learning can be incorporated into adversarial training and the choices to consider: (a) self-supervision task, (b) generation of adversarial examples, and (c) network parameter optimization. We find that leveraging the self-supervised task loss in generating adversarial examples (loss $\mathcal{L}$ at the bottom of Figure \ref{fig:diagram}) is crucial to boosting model robustness, however it penalizes its non-robust accuracy\footnote{This is expected, as there is much work exploring the trade-off between robustness and accuracy \cite{tsipras2018robustness}.}. 

\textbf{Contributions} In this paper, we conduct a comprehensive study to understand various ways one can combine self-supervision with adversarial training and observe the effects on $l_2$ and $l_{\infty}$ adversarial robustness. Our study is motivated by the following three questions:

\begin{itemize}
    \item Is there a best way to combine self-supervised learning with adversarial training?
    \item Is there anything special about predicting image rotations? Or would other SS tasks work as well.
    \item Does adversarial self-supervised pre-training further boost model robustness if one incorporates self-supervision into adversarial training?
\end{itemize}

Our contributions are the answers to the above questions:
\begin{itemize}
    \item Depends on the meaning of best. If one is interested in model accuracy, incorporating self-supervision loss into optimizing network parameters improves the non-robust model accuracy and robustness for small values $\epsilon$, however it can hurt model robustness for larger perturbations. Optimizing the model with self-supervision and adversarial examples generated with both supervised and self-supervised losses is optimal for boosting model robustness, is this can be viewed as a form ensemble adversarial training.
    \item Our results indicate that there is nothing special about predicting image rotations and its ability to boost model robustness. We see a similar boost in adversarial robustness when swapping image rotations with the task of solving a jigsaw puzzle \cite{carlucci2019domain}.
    \item Our results indicate the answer is no. Self-supervised adversarial pre-training does boost accuracy if starting adversarial training \cite{madry2018towards} from pre-trained weights. However we see no further benefit from SS adversarial pre-training if we incorporate SS into adversarial training.
\end{itemize}

\section{Related work}
\subsection{Self-supervised learning}
Self-supervised learning has gained popularity in recent years in many domains as a method of learning  feature representations without the need of labeled data, e.g. images \cite{rotnet}, or video \cite{Xu_2019_CVPR}. Many of the visual self-supervised learning methods utilize a particular structure of the data to create a task and generate pseudo-labels that can then be optimized using supervised learning, e.g image colorization \cite{zhang2016colorful}, or rotations prediction \cite{rotnet}. Noroozi et al. \cite{jigsaw} propose to learn features by solving jigsaw puzzles on natural images. Zhang et al. \cite{zhang2016colorful} propose to learn image features by learning to predict plausible color version of a photograph from its greyscale version. Gidaris et al \cite{rotnet} propose to learn visual features by predicting the nature of geometrical transformation applied to the image, in this case image rotations. Trinh et al. \cite{selfie} learn visual features by learning to predict which image patches correctly fill in masked out image among a set of ``distractor" patches. Further, some of the recent methods such as SwAV \cite{swav} or MoCo-V2 \cite{mocov2} utilize contrastive learning and feature clustering to learning strong feature representations.

\subsection{Adversarial robustness}
Though neural networks have been in the forefront of state-of-the-art research in many branches of computer vision, they exhibit some weaknesses, among them vulnerability to adversarial examples \cite{ szegedy2013intriguing, goodfellow2015advexamples}, which completely change the output of the model. Recent years have seen many attempts to propose defenses against such examples. Dhillon et al. \cite{stochasticDefense} propose Stochastic Activation Pruning, a strategy which randomly prunes a subset of activations. Guo et al. \cite{guo2018countering} propose to use input transformation (e.g. JPEG compression) to defend against adversarial attacks. Xie et al. \cite{xie2018mitigating} attempt to defend against adversarial attacks by adding an additional layer that randomly rescales the input. However, as Athalye et al. \cite{falseSense} show, many of these defenses can be easily circumvented. One of the most effective approaches to defending against adversarial attacks is shown to be adversarial training \cite{madry2018towards, falseSense}. 

\section{Improving robustness with self-supervision}
In this section, we describe the form of our empirical study into the role of self-supervision and its effects on model robustness. Previous studies, most notably Chen et al. \cite{sslRobustness} and Hendrycks et al. \cite{sslHendrycks}, explore using self-supervision to improve model robustness via pre-training and as an auxiliary loss, both focusing on the particular setting where the adversarial examples are generated with projected gradient descent (PGD) attacks \cite{madry2018towards} with the $l_{\infty}$ norm and attack strength equal to $\epsilon = \frac{8}{255}$. A recent empirical study by Salman et al. \cite{robustTransfer} shows that model robustness (mainly models trained with small values of $l_2$ perturbation) can improve the performance in downstream learning tasks. Motivated by this research, our empirical study focuses on understanding the effect of self-supervision on model robustness in case of both $l_2$ and $l_{\infty}$ adversarial perturbations of various strength.

\subsection{Self-supervised learning}
As LeCun et al. \cite{darkMatter} point out, ``self-supervised learning obtains supervisory signals from the data itself, often leveraging the underlying structure in the data." Imagine we are given an input image $x$ and we apply to it a transformation $T(\cdot)$ that gives us the a transformed version of the image $x_{SS}$ and label $y_{SS}$:
\begin{equation}
    (x_{SS}, y_{SS}) = T(x).
\end{equation}
In our experiments, we study the following self-supervised task: rotation prediction \cite{rotnet}, and solving a jigsaw puzzle \cite{jigsaw, carlucci2019domain} (see Figure \ref{fig:diagram}). Both of these tasks are implemented as a supervised task with multi-class cross-entropy loss as a criterion to be optimized. This allows us to easily incorporate rotation prediction as a auxiliary task to be optimized in addition to classification. In case of rotations predictions, the transformation $T(\cdot)$ consists of rotating the image by a multiple of $90\deg$, and returning a rotated image $x_{SS}$ and a label $y_{SS}$ assigned to the particular rotation. In case of the jigsaw puzzle, the task of consists of dividing the image into $n \times n$ grid, and randomly picking a set of $J$ permutations (each assigned its own label). The transformation $T(\cdot)$ then consists of scrambling the image parts according to the randomly chosen permutation, returning the scrambled image and label associated with the permutation. The task of the network is then to either predict the rotation of the image, or the permutation that would unscramble the image.

\subsection{Defending against adversarial perturbations}
To set the setting, we will borrow the notation from Madry et al. \cite{madry2018towards, tsipras2018robustness}. Typically, machine learning models are optimized to maximize a certain metric and thus have a low expected loss,
\begin{equation}
    \underset{(x, y) \sim \mathcal{D}}{\EX}  \left[\mathcal{L}(x, y; \theta) \right].
\end{equation}
As mentioned earlier, neural networks were found to be vulnerable to adversarial examples \cite{szegedy2013intriguing, goodfellow15}, which are images that have been altered to fool a model. Therefore, there is a lot of interest in developing models that are \say{robust}, i.e. resistant to adversarial examples. Therefore, the goal is now to train models that minimize the adversarial loss,
\begin{equation}
    \underset{(x, y) \sim \mathcal{D}}{\EX}  \left[ \max_{\delta \in \Delta} \mathcal{L}(x+\delta, y; \theta) \right],
\end{equation}
where $\Delta$ is a set of $l_p$-bounded perturbations, or alternatively a set $S$ of adversarial examples $x' \in \mathcal{X}$ where $S = \{ x' \in \mathcal{X} : \lVert x' - x \rVert_p < \epsilon \}$.

One of the most successful methods for defending is adversarial training \cite{goodfellow15, madry2018towards, wong2018provable}. A variant of adversarial training proposed by Madry et al \cite{madry2018towards} solves the above problem by finding the worst case $l_p$ bounded adversarial examples using projected gradient descent (PGD) and finds set of parameters $\theta$ that minimize the empirical training loss using these examples \cite{falseSense},
\begin{equation}
    \theta^* = \underset{\theta}{\arg \min} \underset{(x, y) \in X}{\EX} \left[ \underset{\delta \in [ -\epsilon, \epsilon]}{\max} \mathcal{L}(x+\delta, y; F_{\theta})  \right] 
\end{equation}

\begin{algorithm}[t]
\caption{Finding adversaries with self-supervision}\label{alg:pgd}
\begin{algorithmic}[1]
\State \textbf{input}: Robustness parameter $\epsilon$, attack learning rate $\alpha$, number of attack steps $K$, image batches $X$ and $X_{SS}$, labels $y$ and $y_{SS}$, $\lambda_2$ 
\State \textbf{output}: Adversarial images $X$ and optionally $X_{SS}$
\State Randomly perturb images $X$ (and optionally $X_{SS}$)
\State $X^0 = X + U(-\epsilon, \epsilon), X^0_{SS} = X_{SS} + U(-\epsilon, \epsilon)$
\For{$k=0, \dots, K-1$}
    \State $\mathcal{L} = \mathcal{L}_{sup}(F_{\theta}(X^{k}), y)$
    \If{should use self-supervised loss}
        \State $\mathcal{L} \mathrel{+}= \lambda_2 * \mathcal{L}_{SS}( F_{\theta}^{SS}(X_{SS}^{k}), y_{SS})$
    \EndIf
\EndFor
\State $X^{k+1} = \Pi_S \left( X^k + \alpha \cdot \text{sign} \left( \nabla_x \mathcal{L} \right) \right)$
\end{algorithmic}
\end{algorithm}

In adversarial training by Madry et al. \cite{madry2018towards}, an image $x$ is first perturbed in its neighborhood $U(x, \epsilon)$, i.e.
\begin{equation}
    x_0 = x + U(-\epsilon, \epsilon).
\end{equation}
It then follows the generation process of the Basic Iterative Method \cite{kurakin2016adversarial, pang2019improving}, 
\begin{equation}
    x^{k+1} = \Pi_S \left( x_k + \alpha \cdot \text{sign} \left( \nabla_x \mathcal{L}(x+\delta, y; F_{\theta}) \right) \right) 
\end{equation}
where $K$ is the number of \say{attacks steps}, or iterations of projected gradient descent performed to find the worst case $l_p$ bounded adversarial example $x^{adv}$. $\Pi_p$ is an operator that will project the iterates onto an $l_p$ ball, in our case $p = 2$ and $\infty$.

Algorithm \ref{alg:pgd} describes the general procedure for generating PGD adversaries. As we see in Figure \ref{fig:diagram} and Algorithm \ref{alg:pgd}, there are several choices to be made regarding how to incorporate self-supervision into adversarial training (AT). For now, assume our self-supervised task of choice is rotation prediction \cite{rotnet}, though we show that there is nothing unique about rotation prediction and one can substitute any self-supervised task of choice. One can use image rotation in the image pre-processing pipeline (without incorporating self-supervised loss), however this hurts the baseline AT performance, and therefore is not further considered.  

The first way to incorporate self-supervision into AT is to apply the transformation $T$ to the batch of images $X$ to obtain the rotated versions of the batch and corresponding labels $X_{SS}$ and $y_{SS}$ respectively. One then computes the adversarial version of $X_{SS}^{adv}$ and compute the following loss 
\begin{equation}
    \mathcal{L}_{sup}(F_{\theta}(X_{SS}^{adv}), y) + \lambda_1 * \mathcal{L}_{SS}(F_{\theta}^{SS}(X_{SS}^{adv}), y_{SS}),
\end{equation}
where $\lambda_1$ is the weight of the self-supervised loss\footnote{Note that the loss $\mathcal{L}$ also includes the weight regularization term. however this is not shown as this term is present for all of the cases.}, $F_{\theta}$ represents a neural network model with two prediction heads, one for the supervised task ($F_{\theta}(\cdot)$) and the other for self-supervised task ($F_{\theta}^{SS}(\cdot)$). This too is found to be inferior to baseline adversarial training, dropping the clean and robust ($\epsilon_{test} = 8/255$) accuracy from 85.46\% and 44.69\% to 74.9\% and 36.60\% respectively.

\begin{figure*}[t]
  \centering
    \includegraphics[width=0.95\linewidth]{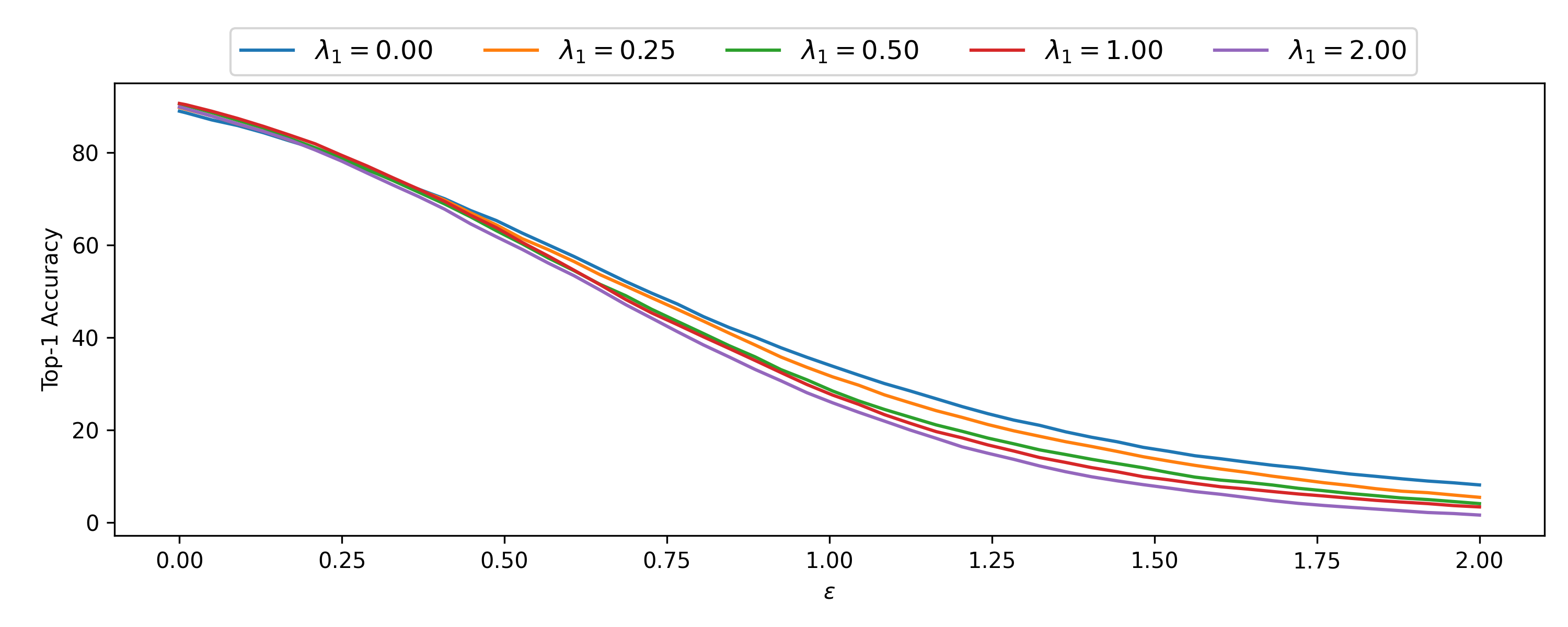}
   
    \includegraphics[width=0.33\linewidth]{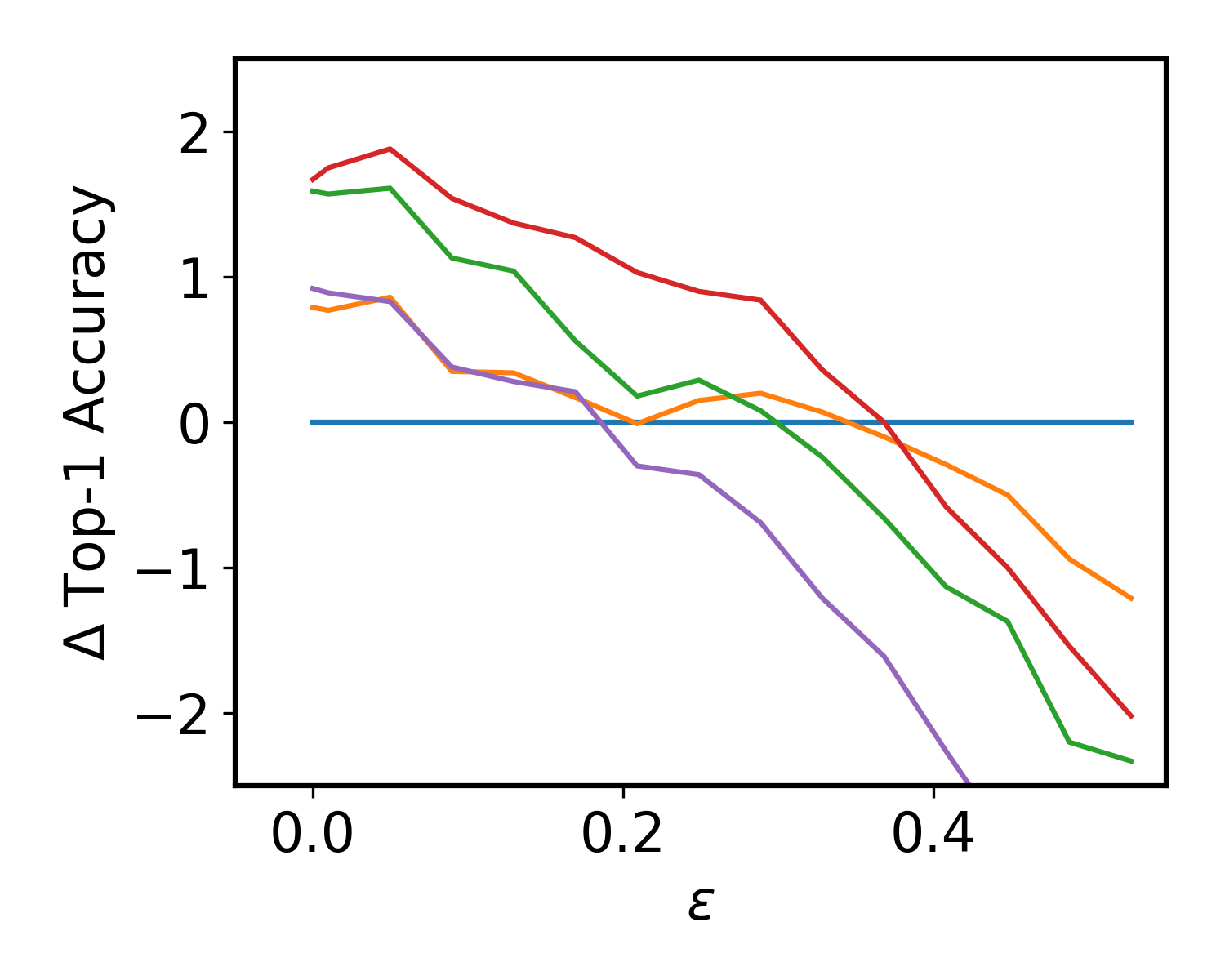}
    \includegraphics[width=0.33\linewidth]{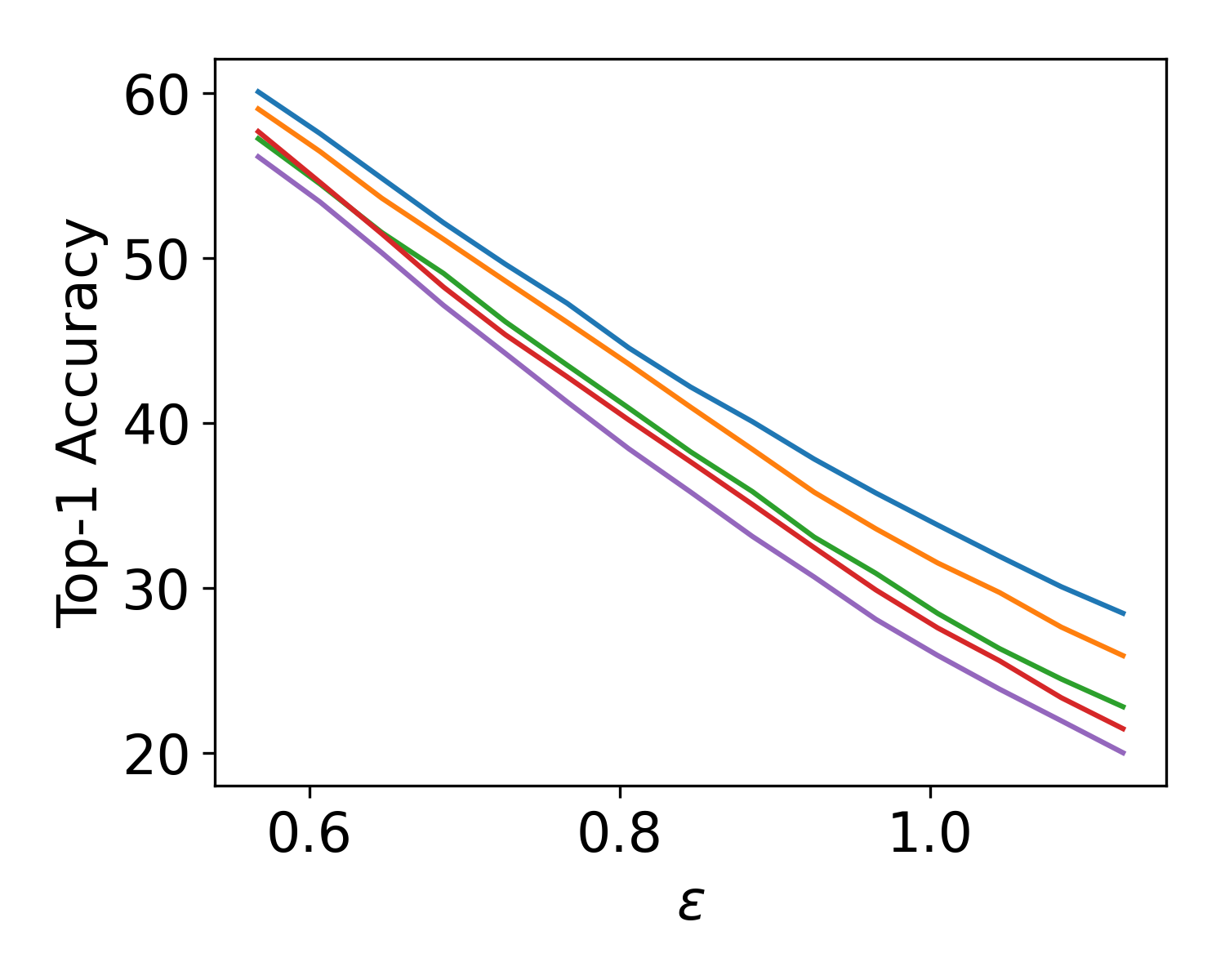}
    \includegraphics[width=0.33\linewidth]{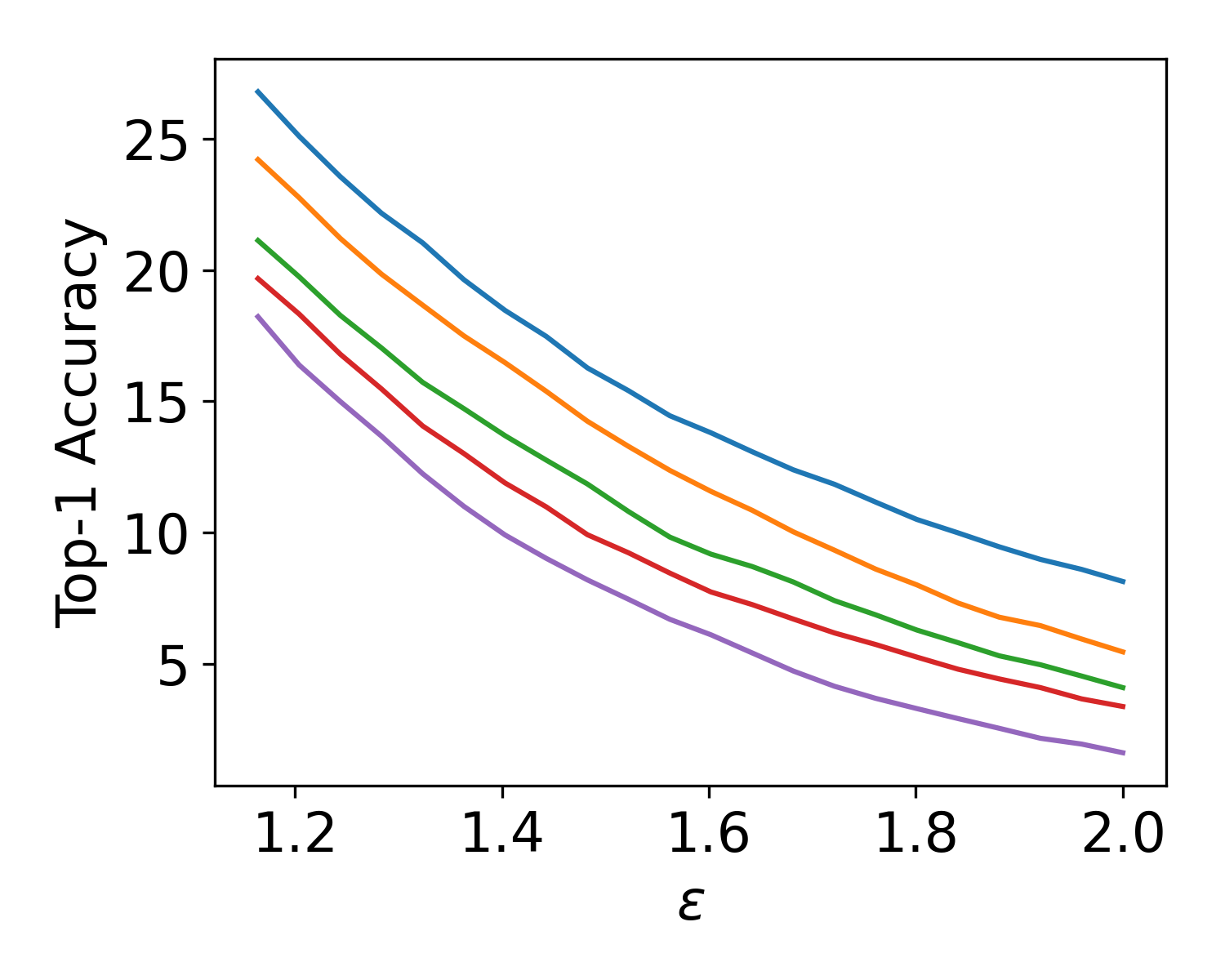}

   \caption{A comparison of models for which $\lambda_1$ is varied. The top figure shows the full plot for values of $\epsilon_{test}$ The bottom row shows figures highlighted different parts of the domain. Leftmost plot rather than showing the performance of the various models, shows the comparison of performance for the given model, e.g. the line in the bottom left plot that shares the yellow colour with the plot in the top figure, shows the difference in performance $\text{Acc}(\lambda_1 = 0.25) - \text{Acc}(\lambda_1 = 0)$.}
   \label{fig:sslv1}
\end{figure*}

The following are options for adding self-supervision that involve making a copy of the batch to which the self-supervised transformation $T$ will be applied, resulting in a batch $X$ and a batch $X_{SS} = T(X)$\footnote{Transformation $T$ also gives us the self-supervised labels $y_{SS}$}. These scenarios fit the description of Algorithm \ref{alg:pgd} and Figure \ref{fig:diagram}. The simplest way to add self-supervision into AT (denoted as SS $\mathcal{T}_1$) is by generating adversarial examples $X^{adv}$, while leaving $X_{SS}$ unchanged, and computing the combined loss 
\begin{equation}
    \mathcal{L}_{sup}(F_{\theta}(X^{adv}), y) + \lambda_1 * \mathcal{L}_{SS}(F_{\theta}^{SS}(X_{SS}), y_{SS})\label{eq:ssv1}
\end{equation} Lastly, one can attack both the original batch images $X$ and transformed images $X_{SS}$, resulting in the following loss
\begin{equation}\label{eq:ssv2}
    \mathcal{L}_{sup}(F_{\theta}(X^{adv}), y) + \lambda_1 * \mathcal{L}_{SS}(F_{\theta}^{SS}(X_{SS}^{adv}), y_{SS}).
\end{equation} 
This scenario offers another set of decisions corresponding to lines 4-6 of Algorithm \ref{alg:pgd}: when generating adversarial examples for $X$ and $X_{SS}$, should one use the supervised loss (SS $\mathcal{T}_2$), the self-supervised loss, or both (and what ratio $\lambda_2$) (SS $\mathcal{T}_3$) to find the worst case adversarial examples? As can be seen, incorporating self-supervision into AT is not straightforward. Using just SS loss for generating adversarial examples is much worse as compared to the baseline, due to the fact that they are not optimized to defend against perturbations that they are tested against.

\subsection{Self-supervised pre-training}
We first study how incorporating SSL into adversarial training impact the model robustness. However, from Chen et al. \cite{sslRobustness} we know that adversarial self-supervised pre-training can help robustness. Therefore, we want to understand whether self-supervised pre-training will further aid in model robustness, or whether majority of the benefits are realized by combining SSL and AT. As Chen et al. point out \cite{sslRobustness}, partial fine-tuning, where a portion of the early layers is frozen, is inferior to tuning the full model (as expected, as it is similar to fixed-feature and full-network transfer learning setting of \cite{robustTransfer}). Therefore, once we have understood the best way to incorporate SSL into AT, we will consider the following settings for initializing our weights: random initialization, adversarial self-supervised pre-training $\mathcal{P}_1$ with $l_2$ perturbations, and adversarial self-supervised pre-training $\mathcal{P}_2$ with $l_{\infty}$. This is equivalent to AT, except $X_{SS}$ and $y_{SS}$ are used for AT algorithm of \cite{madry2018towards}.

\begin{table*}[t]
\begin{center}
\resizebox{1.0\textwidth}{!}{%
\begin{tabular}{lrrrrrrrrrrrr}
\toprule
Method / $\epsilon_{test}$ &      0 &   0.01 &   0.03 &   0.05 &   0.07 &    0.1 &   0.25 &    0.5 &   0.75 &    1.0 &    2.0 &   3.0 \\
\midrule
Baseline $\mathcal{T}_0$    &  88.98 &  88.63 &  87.87 &  87.08 &  86.48 &  85.49 &  78.52 &  64.46 &  48.15 &  34.05 &  8.13 &  2.15 \\
SS $\mathcal{T}_1$          &  90.57 &  90.20 &  89.48 &  88.69 &  87.76 &  86.68 &  78.78 &  62.17 &  44.65 &  28.80 &  4.09 &  0.48 \\
SS $\mathcal{T}_2$          &  90.20 &  89.94 &  89.30 &  88.54 &  87.94 &  87.02 &  80.31 &  66.18 &  50.28 &  34.91 &  6.28 &  1.77 \\
SS $\mathcal{T}_3$          &  89.07 &  88.75 &  88.14 &  87.48 &  86.85 &  85.73 &  79.77 &  66.15 &  50.87 &  36.55 &  7.50 &  2.09 \\
\bottomrule

\end{tabular}
}
\vspace{-1em}
\end{center}
\caption{ Top-1 accuracy on the CIFAR-10 test set with ResNet-18 models trained with $l_{2}$ norm and $\epsilon_{train} = 0.5$ evaluated at different values of $\epsilon_{test}$ used to generate the robust test set as we vary the training method and SSL involvement.
}
\label{tab:rn18_l2_1}
\end{table*}

\begin{table*}[]
\begin{center}
\resizebox{0.95\textwidth}{!}{%
\begin{tabular}{lrrrrrrrrr}
\toprule
Method / $\epsilon_{test}$ &  $0/255$ &  $3/255$ &  $4/255$ &  $5/255$ &  $6/255$ &  $7/255$ &  $8/255$ &  $9/255$ &  $10/255$ \\
\midrule
Baseline $\mathcal{T}_0$    &    85.46 &    72.09 &    66.98 &    61.29 &    55.83 &    50.19 &    44.69 &    39.38 &     34.60 \\
SS $\mathcal{T}_1$          &    86.59 &    73.69 &    68.71 &    62.86 &    56.88 &    50.52 &    44.51 &    38.90 &     33.89 \\
SS $\mathcal{T}_2$          &    86.24 &    74.06 &    68.96 &    63.29 &    57.63 &    51.50 &    45.77 &    40.51 &     35.54 \\
SS $\mathcal{T}_3$          &    84.61 &    73.33 &    68.61 &    64.01 &    59.09 &    53.73 &    48.11 &    43.42 &     38.76 \\
\bottomrule
\end{tabular}
}
\vspace{-1em}
\end{center}
\caption{ Top-1 accuracy on the CIFAR-10 test set with ResNet-18 models trained with $l_{\infty}$ norm and $\epsilon_{train} = 8/255$ evaluated at different values of $\epsilon_{test}$ used to generate the robust test set as we vary the training method and SSL involvement.
}
\label{tab:rn18_linf_1}
\end{table*}

\section{Experiments and results}

\subsection{Datasets}
Our experiments consider the following datasets: CIFAR-10, and CIFAR-10C \cite{hendrycks2018benchmarking}. All of the ablation experiments are done using CIFAR-10. The CIFAR-10C dataset \cite{hendrycks2018benchmarking} is used to test our model against common corruptions that can be encountered.

\subsection{Implementation details}
All of our code is implemented in Python using the PyTorch \cite{PyTorch} deep learning framework. In pursuit of simplicity and reproducibility, our code extends the \textit{robustness} library \cite{robustness}\footnote{Source code can be made available upon internal review.}.  Our code is developed on an internal cluster, where each server node is equipped with 4 NVIDIA Tesla P100 cards (each with 16 GB of VRAM), paired with a dual 18-core Intel Xeon CPUs and 256GB of memory. Our ablation experiments utilize the ResNet-18, and ResNet-34 architectures\footnote{Note that the robustness package uses slightly different architectures for the CIFAR-10 and ImageNet dataset. CIFAR-10 uses slightly smaller convolutional kernels. For more details, please see: \hyperlink{robustess}{https://github.com/MadryLab/robustness}}. 

When applying self-supervised transformation $T(\cdot)$ to $X$, each image is randomly rotated by $\phi$ degrees, where $\phi \in \{0^{\circ}, 90^{\circ}, 180^{\circ}, 270^{\circ} \}$. This implementation differs from \cite{sslHendrycks}, as they simultaneously generate and predict all four possible rotations for each image in the batch (if the size of $X$ is $b$ samples, $X_{SS}$ is of size $4 \cdot b$). Furthermore, when leveraging SS loss to guide adversarial training (step 6 of Algorithm \ref{alg:pgd}), we use the following form of the loss:
\begin{equation}
    \mathcal{L}_{SS}(X_{SS}, y_{SS}) = \frac{1}{N}\sum_{i \in \{1, \cdots, n \}} \mathcal{L}_{CE}(F^{SS}_{\theta}(x^i_{SS}), y^i_{SS}),
\end{equation}
which differs from \cite{sslHendrycks} in that they omit the averaging term $1/N$. In our experiments, averaging reduction for the loss gives consistently better results.

\textbf{Training and evaluation details} All of the ablation experiments investigating addition of SS into AT are trained for 100 epochs, with a starting learning rate of 0.1 (reduced by 10 every 40 epochs) and optimized using Stochastic Gradient Descent (SGD) with the momentum of 0.9. The CIFAR-10 training set is further split into a training and validation set (15\% or 7500 images are used for validation and the rest are used for training). For adversarial training and evaluation, we use 10 and 20-step $l_p$ PGD attacks \cite{madry2018towards} respectively. For $l_2$ adversaries, the attack learning rate is set to $\alpha = \frac{\epsilon \cdot 2}{\# \text{ of steps}}$ (see supplement of \cite{robustTransfer}). For $l_{\infty}$ adversaries, the $\epsilon_{train} = 8/255$ and attack learning rate is set to $\alpha=2/255$ \cite{sslHendrycks, sslRobustness}. Similar to \cite{sslRobustness}, in our results we refer to the non-robust, i.e. $\epsilon=0$, accuracy as the \textit{Standard Testing Accuracy} (TA), and robust accuracy at $\epsilon_{test} = \epsilon$ as the \textit{Robust Testing Accuracy} ($RA_{\epsilon}$). For particular set of hyper-parameters, we pick a model with the highest validation TA and test across a range of $\epsilon_{test}$ values. The adversarial perturbations for testing images are only generating using the supervision loss in the PGD attack.

\begin{table*}[t]
\begin{center}
\resizebox{0.95\textwidth}{!}{%
\begin{tabular}{lrrrrrrrrr}
\toprule
Method / $\epsilon_{test}$ &  $0/255$ &  $3/255$ &  $4/255$ &  $5/255$ &  $6/255$ &  $7/255$ &  $8/255$ &  $9/255$ &  $10/255$ \\
\midrule
Baseline    &    85.46 &    72.09 &    66.98 &    61.29 &    55.83 &    50.19 &    44.69 &    39.38 &     34.60 \\
Rotations   &    84.61 &    73.33 &    68.61 &    64.01 &    59.09 &    53.73 &    48.11 &    43.42 &     38.76 \\
Jigsaw      &    83.90 &    72.39 &    67.91 &    63.18 &    57.99 &    52.89 &    47.57 &    42.31 &     37.66 \\
\bottomrule
\end{tabular}
}
\vspace{-1em}
\end{center}
\caption{ Top-1 accuracy on the CIFAR-10 test set with ResNet-18 models trained with $l_{\infty}$ norm and $\epsilon_{train} = 8/255$ evaluated at different values of $\epsilon_{test}$ used to generate the robust test set as we vary the SSL tasks that is used in tandem with AT.
}
\label{tab:linf_rn18_tasks}
\end{table*}

\subsection{Effect of self-supervision on model robustness}\label{sec:at}

In this section, we compare the ways SS can be incorporated into AT (SS $\mathcal{T}_1$, SS $\mathcal{T}_2$, and SS $\mathcal{T}_3$) and we compare it against the baseline AT procedure of Madry et al. \cite{madry2018towards} ( Baseline $\mathcal{T}_0$). Tables \ref{tab:rn18_l2_1} and \ref{tab:rn18_linf_1} summarize the general trends we see for the $l_2$ and $l_\infty$-robust performance of the various methods for the ResNet18 architecture. We see similar trends for the ResNet-34 architecture, whose tabulated results can be found in the appendix.

Our first set of experiments explores the effects on AT of the training scenario SS $\mathcal{T}_1$, which optimizes
\begin{equation}
    \mathcal{L}_{sup}(F_{\theta}(X^{adv}), y) + \lambda_1 * \mathcal{L}_{SS}(F_{\theta}^{SS}(X_{SS}), y_{SS}).
\end{equation}
$\mathcal{L}_{sup}$ is the supervised portion of the loss operating on PGD-attacked batch $X^{adv}$, and $\mathcal{L}_{SS}$ is the self-supervised loss (SSL) operating on clean batch $X_{SS}$. Figure \ref{fig:sslv1} illustrates the general trend we see for SS $\mathcal{T}_1$. The bottom row of Figure \ref{fig:sslv1} splits the $x$-axis into three ranges, to better illustrate the trend. The tabulated results for various values of $\epsilon_{test}$ and $\lambda_1$ can be found in Supplemental Materials. In general, we see that TA is higher for models with $\lambda_1 > 0$. In fact, the TA for models trained under SS $\mathcal{T}_1$ scenario is better than the non-robust model with maximum allowable $l_2$ perturbation $\epsilon_{train}$ as high as 0.1. The use of self-supervision ($\alpha_1 > 0$) consistently improves the TA and RA for several values of $\epsilon_{test} < \epsilon_{train}$, as compared to baseline $\mathcal{T}_0$ (see bottom left plot of Figure \ref{fig:sslv1}). Furthermore, larger values of $\alpha_1$ result in larger increase in TA and RA for several values of $\epsilon_{test} < \epsilon_{train}$. However, for $\epsilon_{test} > \epsilon_{train}$ we can see the opposite behavior - RA drops further as compared to the baseline the larger the weight $\lambda_1$. From Figure \ref{fig:sslv1}, and Tables \ref{tab:rn18_l2_1} and \ref{tab:rn18_linf_1}, we can see that SS $\mathcal{T}_1$ is very effective in regularizing the model training and provides the largest increase in TA as compared to other methods (though sacrificing RA).

\begin{figure*}[]
  \centering
   \includegraphics[width=0.98\linewidth]{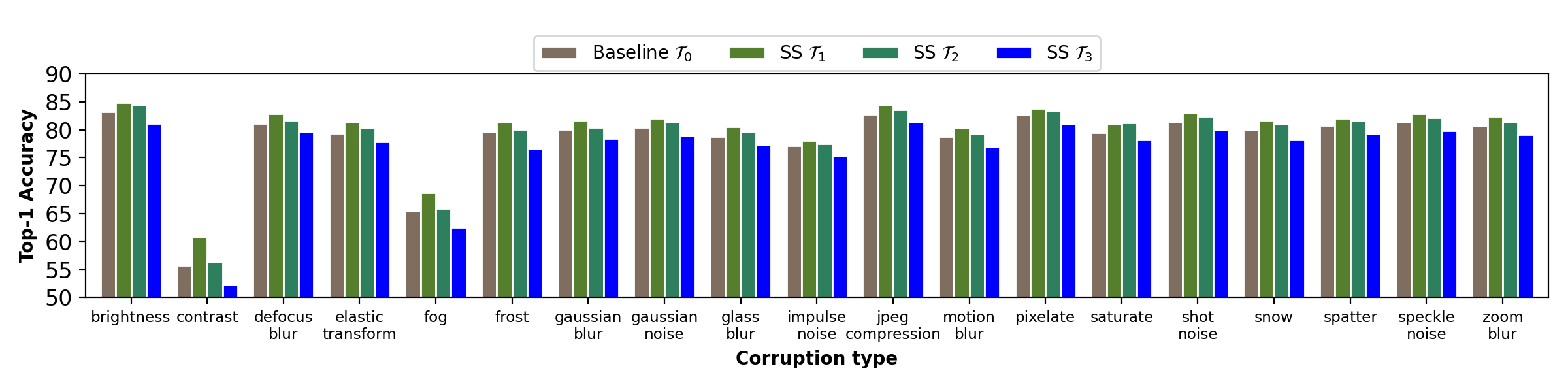}
   \includegraphics[width=0.98\linewidth]{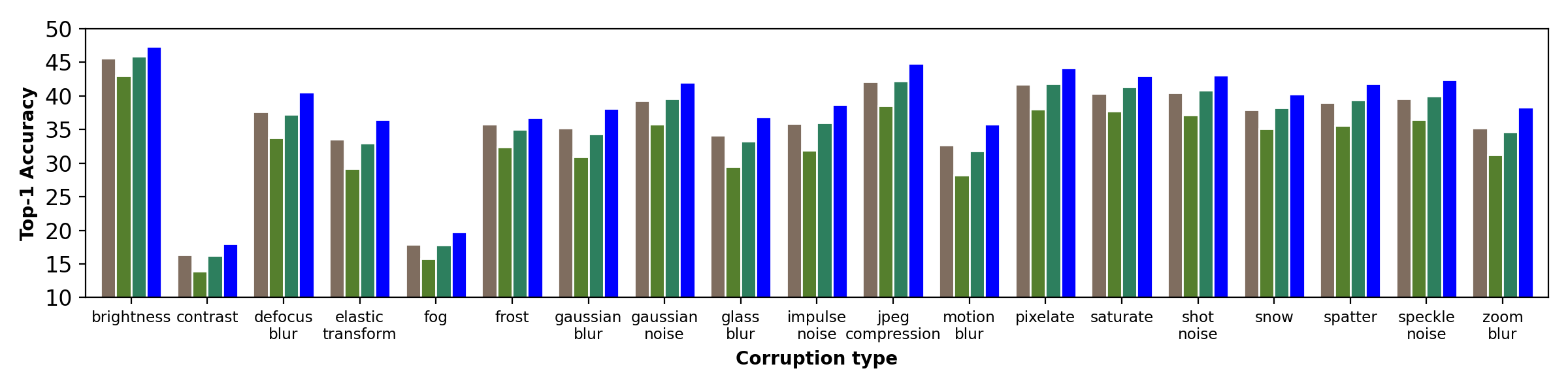}
   \caption{Figure depicting the Top-1 accuracy comparing the various methods to incorporate self-supervision into adversarial training for the various corruption types present in the corrupted CIFAR-10 dataset \cite{hendrycks2018benchmarking}. Top figure shows the non-robust Top-1 accuracy and bottom figure shows the robust Top-1 accuracy. The baseline ResNet-18 model is trained with a maximum allowable $l_2$ corruptions of $\epsilon=1.0$. }
   \label{fig:cifar10c}
\end{figure*}

\begin{table*}[t]
\begin{center}
\resizebox{0.95\textwidth}{!}{%
\begin{tabular}{lrrrrrrrrr}
\toprule
Method / $\epsilon_{test}$ &  $0/255$ &  $3/255$ &  $4/255$ &  $5/255$ &  $6/255$ &  $7/255$ &  $8/255$ &  $9/255$ &  $10/255$ \\
\midrule
Baseline $\mathcal{P}_0 | \mathcal{T}_0$    &    85.29 &    71.90 &    66.63 &    60.96 &    55.42 &    49.52 &    44.01 &    39.28 &     34.41 \\ \hline
$\mathcal{P}_1 | \mathcal{T}_0$	            &    84.67 &    72.67 &    67.91 &    62.61 &    56.91 &    51.46 &    45.91 &    40.27 &     35.88 \\
$\mathcal{P}_2 | \mathcal{T}_0$             &    85.37 &    72.01 &    66.69 &    61.05 &    55.52 &    49.61 &    43.70 &    38.17 &     33.66 \\
$\mathcal{P}_1 | \mathcal{T}_3$             &    83.78 &    72.75 &    68.20 &    63.32 &    58.25 &    52.85 &    47.37 &    42.24 &     37.93 \\
$\mathcal{P}_2 | \mathcal{T}_3$             &    84.08 &    72.65 &    68.08 &    62.94 &    58.01 &    52.77 &    47.16 &    41.67 &     36.58 \\ \hline
$\mathcal{P}_0 | \mathcal{T}_3$             &    84.61 &    73.33 &    68.61 &    64.01 &    59.09 &    53.73 &    48.11 &    43.42 &     38.76 \\
\bottomrule
\end{tabular}
}
\vspace{-1em}
\end{center}
\caption{ Top-1 accuracy on the CIFAR-10 test set with ResNet-18 models trained with different combinations of pre-training and AT.
}
\label{tab:linf_rn18_pretraining}
\end{table*}

The row SS $\mathcal{T}_2$ and The row SS $\mathcal{T}_3$ from tables \ref{tab:rn18_l2_1} and \ref{tab:rn18_linf_1} show the cases in which both original ($X$) and transformed images ($X_{SS}$) were attacked using the PGD (Alg. \ref{alg:pgd}) and use the following loss for network parameter optimization: 
\begin{equation}
    \mathcal{L}_{sup}(F_{\theta}(X^{adv}), y) + \lambda_1 * \mathcal{L}_{SS}(F_{\theta}^{SS}(X_{SS}^{adv}), y_{SS}).
\end{equation}
The main difference between rows SS $\mathcal{T}_2$ and SS $\mathcal{T}_3$ is the manner in which the adversarial images are generated. Row SS $\mathcal{T}_2$ shows result for a ResNet-18 trained model in which only line 6 of Algorithm \ref{alg:pgd} is used to generate loss for the PGD attack, whereas row SS $\mathcal{T}_3$ shows the results for the case in which line 8 of Algorithm \ref{alg:pgd} is used to add SSL term. For the case of $\mathcal{T}_2$, in both $l_2$ (Table \ref{tab:rn18_l2_1}) and $l_{\infty}$ (Table \ref{tab:rn18_linf_1}) we can see that the TA and RA are better than $\mathcal{T}_0$ for all values of $\epsilon_{test}$, though TA is worse than SS $\mathcal{T}_1$. Lastly, for SS $\mathcal{T}_3$ we can see that TA either does not change or gets worse compared to all the other methods, showing that if one wants to optimize the model for clean accuracy, one should not use SSL in generating adversarial examples. However, where $\mathcal{T}_3$ shines is its increase in RA as compared to baseline or the other methods, especially for larger values of $\epsilon_{test}$. What causes this behavior (drop in RA and increase in TA)? By attacking both the supervision and self-supervision loss, we are optimizing the model to be robust to multiple objectives, which can be viewed as a form of ensemble adversarial training \cite{tramer2020ensemble, beyondAT}, in which one performs an ensemble attack over multiple tasks \cite{wang2021adversarial}.

\textbf{Discussion} From our empirical study, we can see that both (a) combining the supervision with self-supervised loss, and (b) attacking both the supervision and self-supervision losses play a vital role in affecting the TA and robustness to $l_2$ and $l_{\infty}$-PGD attacks, which is contrary to findings of Hendrycks et al. \cite{sslRobustness}. Just adding SS loss improves the TA and RA for $\epsilon_{test} \le \epsilon_{train}$, however hurts in for $\epsilon_{test} > \epsilon_{train}$. Using the adversarial version of both $X$ and $X_{SS}$ (using only the supervised loss) slightly improves the accuracy for all $\epsilon_{test}$ as compared to $\mathcal{T}_1$, however slightly lowers TA and RA for small values of $\epsilon$. SS $\mathcal{T}_3$ can be viewed as a form of a ensemble adversarial attack over multiple tasks, resulting in an increase in model robustness to larger perturbations however sacrificing TA \cite{tsipras2018robustness}.  

\textbf{Is there something special about rotation prediction?} Our experiments suggest the answer to be no. Refer to Table~\ref{tab:linf_rn18_tasks}, where we compare swapping the task of predicting image rotations to the jigsaw task \cite{carlucci2019domain}, where one scrambles the image parts into one of several pre-selected permutations and then predicts the nature of the permutation. We can see from Table~\ref{tab:linf_rn18_tasks} that either the rotations or jigsaw task succeed in improving the robustness of the model in a similar way, though we can see that jigsaw is slightly inferior to rotation prediction. This is similar to the result of \cite{sslRobustness}, which observes that jigsaw pre-training provides worse pre-training performance as compared to rotations. 

\textbf{How beneficial is pre-training?} Another question is: how beneficial is self-supervised adversarial pre-training to SS $\mathcal{T}_3$. Please see Table \ref{tab:linf_rn18_pretraining} for a comparison of the training methods with the different pre-training scenarios. Similar to \cite{sslRobustness}, we see a boost in model robustness when adversarial self-supervised pre-training is followed by AT. However, we do not see any significant benefit from adversarial SS pre-training if AT is combined with self-supervision (SS $\mathcal{T}_3$).

\textbf{Clarification of contributions} We provide a comprehensive study on the effects of SS and training choices on model robustness, with explanations of observed trends. Although this study is limited to one dataset and two architectures, this work is a solid platform for exploring and understanding the ways that incorporating SS into AT affects model accuracy and robustness; with findings that reach significantly beyond results from other published work.

\subsection {Robustness to natural corruptions}
In this section, we report the performance of the various SS training on the CIFAR-10C dataset, which contains various common corruptions, e.g. Gaussian noise, blur. Figure \ref{fig:cifar10c} shows the plots of the non-robust (Top) and robust (Bottom) Top-1 accuracies for $l_2$-robust ResNet18 model trained with $\epsilon = 1.0$ (the plot for $l_{\infty}$-robust model can be found in the supplement). From all of these examples we can see the following trend (which is similar to the $l_{\infty}$ trained models): adding the self-supervised loss (SS $\mathcal{T}_1$ and $\mathcal{T}_2$) helps the model improve its TA accuracy for the various common image corruptions - the clean accuracy goes from 78.36\% mean accuracy over all corruptions to  80.23\% and 79.15\% for $\mathcal{T}_1$ and $\mathcal{T}_2$ trained models respectively. The clean accuracy drops from  78.36\% mean accuracy to 76.52\% when SS task loss is added to PGD. However, one can see the benefit of $\mathcal{T}_3$ when adversarial perturbations are added to natural corruptions as RA goes from 35.83\% to 38.33\% mean RA for $l_2$-robust models and from 32.20\% to 36.22\% mean RA for $l_\infty$-robust models. In terms of RA, $\mathcal{T}_1$ and $\mathcal{T}_2$ are worse than $\mathcal{T}_0$ in case of $l_2$-robust models) and  $\mathcal{T}_1$ is worse than  $\mathcal{T}_0$ for $l_{\infty}$. Interestingly, even though the mean TA for both $l_2$ and $l_{\infty}$ robust models is similar, $l_2$-robust models (trained using $\mathcal{T}_0$, $\mathcal{T}_2$, $\mathcal{T}_3)$ outperformed $l_{\infty}$ robust models.

\section{Conclusion}
Our paper aims to understand how incorporating self-supervision into adversarial training improves model robustness to adversarial perturbations and natural corruptions. As we saw, the task of incorporating SS into AT is not straightforward and requires care. Generating a separate image batch transformed by the self-supervised task and incorporating self-supervised task loss into the training objective provides a significant boost in standard testing accuracy, though can hurt the robust testing accuracy for larger values of $\epsilon_{test}$. If the main goal is to improve model robustness, one should should take the approach of ensemble adversarial training (EAT) and generate images that are trying to fool both the self-supervised and supervised training objectives. Though EAT results in a drop in clean accuracy, we see gains in robustness against adversarial perturbations and natural image corruptions. Additionally, we show that these benefits can be had with different SS tasks (rotation prediction vs. jigsaw).

{\small
\bibliographystyle{ieee_fullname}
\bibliography{cvprsubmission}
}

\end{document}